\pdfoutput=1

\documentclass[11pt]{article}

\usepackage{acl}  

\usepackage{times}
\usepackage{latexsym}

\usepackage[T1]{fontenc}

\usepackage[utf8]{inputenc}

\usepackage{microtype}

\usepackage{rotating}

\usepackage{makecell}

\newcommand{\base}{\ensuremath{\mathcal{B}}\xspace}
\newcommand{\targ}{\ensuremath{\mathcal{T}}\xspace}
\newcommand{\rel}{\ensuremath{\mathcal{R}}\xspace}
\newcommand{\simrel}{{\textit{sim}}\xspace}
\newcommand{\simi}{\ensuremath{\textit{sim}^*}\xspace}
\newcommand{\fame}{{\sc Fame}}

\usepackage{xcolor}
\usepackage{multirow} 
\usepackage{caption}
\usepackage{subcaption}
\usepackage[stable]{footmisc}
\usepackage{makecell}

\newcommand{\remove}[1]{}

\usepackage{arydshln}
\usepackage{color}
\usepackage{xspace}
\usepackage{paralist}
\AtBeginDocument{%
  \providecommand\BibTeX{{%
    \normalfont B\kern-0.5em{\scshape i\kern-0.25em b}\kern-0.8em\TeX}}}

\usepackage{amsmath}
\usepackage{mathtools}

\usepackage{algorithm}
\usepackage{algpseudocode} 

\usepackage{tikz}
\usetikzlibrary{positioning,chains,fit,shapes}
\definecolor{blue}{RGB}{80,80,200}
\definecolor{green}{RGB}{80,160,80}
\definecolor{purple}{RGB}{128,0,128}
\definecolor{almostwhite}{RGB}{237,240,235}

\DeclareMathOperator*{\argmax}{arg\,max}
\newcommand\textpaddingnodes{0.2cm}
\newcommand\topbuttonpaddingnodes{2mm}

\newcommand{\dnote}[1]{\textcolor{blue}{$\ll$\textsf{#1 | Dafna}$\gg$}}
\newcommand{\cnote}[1]{\textcolor{red}{$\ll$\textsf{#1 | Chen}$\gg$}}

\newcommand{\xhdr}[1]{\vspace{1mm}\noindent{{\bf #1.}}} 

\usepackage[most]{tcolorbox}
\definecolor{grey}{HTML}{C5C5C5}

\title{FAME: Flexible, Scalable Analogy Mappings Engine}

\author{\vspace{2pt}{Shahar Jacob}, {Chen Shani}, {Dafna Shahaf} \\
\text{The Hebrew University of Jerusalem, Israel} \\
\{shahar.jacob, chenxshani, dshahaf\}@cs.huji.ac.il
}

\begin{document}
\maketitle

\begin{abstract}
    Analogy is one of the core capacities of human cognition; when faced with new situations, we often transfer prior experience from other domains. 
Most work on computational analogy relies heavily on complex, manually crafted input. 
In this work, we relax the input requirements, requiring only names of entities to be mapped. We automatically extract commonsense representations and use them to identify a mapping between the entities. Unlike previous works, our framework can handle partial analogies and suggest new entities to be added. Moreover, our method's output is easily interpretable, allowing for users to understand why a specific mapping was chosen.

Experiments show that our model correctly maps $81.2\%$ of classical 2x2 analogy problems (guess level=$50\%$). On larger problems, it achieves $77.8\%$ accuracy (mean guess level=$13.1\%$). In another experiment, we show our algorithm outperforms human performance, and the automatic suggestions of new entities resemble those suggested by humans. 
We hope this work will advance computational analogy by paving the way to more flexible, realistic input requirements, with broader applicability.

\end{abstract}

\section{Introduction}
\label{sec:intro}
One of the pinnacles of human cognition is the ability to find parallels across distant domains and transfer ideas between them. This \emph{analogous reasoning} process enables us to learn new information faster and solve problems based on prior experience \cite{minsky1988society,hofstadter2013surfaces,holyoak1984analogical,pjm1966models}. 

The most seminal work in computational analogy is Gentner's Structure Mapping Theory (SMT) \cite{gentner1983structure} and its implementation, Structure Mapping Engine (SME) \cite{falkenhainer1989structure}. 
In a nutshell, SMT assumes input from two domains: base and target. It maps between objects in a base domain and objects in a target domain according to common \emph{relational structure}, rather than on object attributes.

For example, consider the Rutherford model of the hydrogen atom, where the atom was explained in terms of the (better-understood) solar system \citep{falkenhainer1989structure}: a planet revolving around the sun is mapped to an electron revolving around the nucleus. The mapping is due to shared \textit{relations} between objects (revolving around, being attracted to), not object attributes (round, small).

\begin{figure*}[h!]
\centering
    \begin{minipage}{.45\textwidth}
        \includegraphics[width=\linewidth]{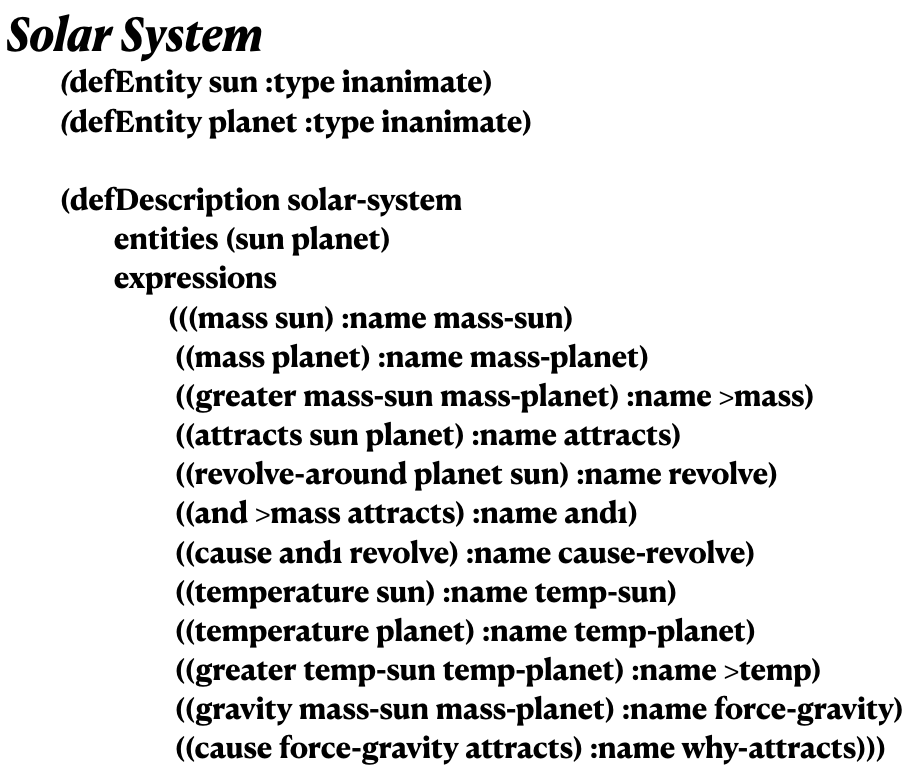}
    \end{minipage}%
    \hfill
    \begin{minipage}{0.45\textwidth}
        \includegraphics[width=\linewidth]{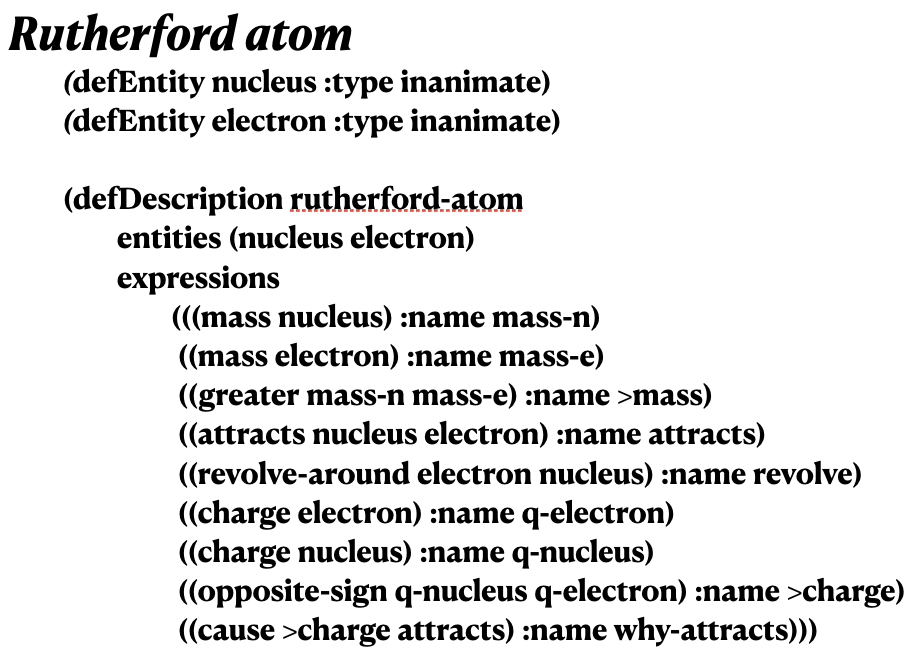}
    \end{minipage}
            \caption{SME representation of the Solar system/Rutherford atom. Reproduced from \citet{falkenhainer1989structure}.}
        \label{fig:SMEatom}
\end{figure*}

One of the main criticisms brought against SME and its follow-up work is their need for extensive hand-coded input -- structured representations of both the entities and their relations (see Figure \ref{fig:SMEatom} for the input to the atom/solar system mapping).

\citet{chalmers1992high} argued that too much human creativity is required to construct this input, and the analogy is already effectively given in the representations: 
``A brief examination [...] shows that the discovery of the similar structure in these representations is not a difficult task. The representations have been set up in such a way that the common structure is immediately apparent. Even for a computer program, the extraction of such common structure is relatively straightforward.''

Some follow-up works avoid hand-coding LISP-like representations, generating them from sketches \cite{forbus2011cogsketch}, qualitative simulators \cite{dehghani2009qcm}, etc. However, they still require much knowledge engineering, and thus are hard to scale.  
Nowadays, when the web is full of information about potential domains to transfer ideas from \cite{mcneil2013car}, such representations do not tap into the potential of web-scale analogies for augmenting human creativity.

The method with the simplest input we are aware of is Latent Relation Mapping Engine (LRME) \cite{turney2008latent}, which requires only two lists of entities to be mapped. Given two entities, they search for phrases containing both in a large corpus and use them to generate simple patterns. For example, ``a sun-centered solar system illustrates'' gives rise to patterns such as ``a X * Y illustrates''.
However, such patterns are extremely simple and brittle, and LRME requires exact string matches between the domains (so ``revolve around'' is different from ``rotate around''). 

In this work, we develop \fame, a Flexible Analogy Mapping Engine. {\fame}'s input requirements are minimal, requiring only two sets of entities. We apply state-of-the-art NLP and IR techniques to automatically infer commonsense relations between the entities using a variety of data sources, and construct a mapping between the domains. Importantly, we do not require identical phrasings of relations. Moreover, our output is interpretable, showing how the mapping was chosen.

Unlike previous works, we drop the strong \emph{bijectivity} assumption and let the algorithm decide which entities to include in the mapping. Meaning, we allow for entities to remain unmapped. Our algorithm can also generate new \emph{suggestions} for the non-mapped entities. This paves the road to algorithms that can handle even more limited input -- for example, using domain \emph{names} (solar system, atom) as input, or just a single mapped entity pairs (e.g., turn white blood cells into policemen and see how the analogy unfolds). 
Our contributions are:

\begin{compactitem}
    \item A novel, scalable, and interpretable approach for automatically mapping two domains based on commonsense \textit{relational} similarities. Our algorithm handles partial mappings and suggests additional entities.
    \item We extend the work of \citet{romero2020inside} to discover salient knowledge about pairs of entities. 
    \item Our model's accuracy is $81.2\%$ on simple, 2x2 problem s(guess level=$50\%$). On larger problems, it achieves $77.8\%$ perfect mappings (guess level=$13.1\%$). In another experiment, we outperform humans ($90\%$ vs.~$70.2\%$) and demonstrate that our automatic suggestions resemble human suggestions. 
     We release code and data.\footnote{\url{https://github.com/shaharjacob/FAME}
        \label{ft:repoLink}
    }
\end{compactitem}

\section{Problem Definition}
\label{sec:problemDefinition}
An analogy is a mapping from a base domain \base into a target domain \targ. The mapping is based on \emph{relations}, and not object attributes. Base objects are not mapped into objects that resemble them; rather, there is a common \textit{relational structure}, and they are mapped to objects that play similar roles. We follow the formulation of \citet{sultan2022life}, brought here for completeness:


\xhdr{Entities and Relations} 
Let \base = \{$b_1, ..., b_n$\} and \targ = \{$t_1, ..., t_m$\} be two sets of entities. For example:
\base = \{sun, Earth, gravity, solar system, Newton\}, 
\targ = \{nucleus, electrons, electricity, atom, Faraday\}.

Let \rel be a set of relations. A relation is a set of ordered entity pairs with some meaning. The exact representation is purposely vague, as we do not restrict ourselves to strings, embeddings, etc. Intuitively, relations should capture notions like ``revolve around''. 

In our example, relations between \base and \targ include the \textit{Earth} revolve around the \textit{Sun}, like \textit{electrons} orbit the \textit{nucleus}; the \textit{Earth} creates a force field of \textit{gravity}, similar to \textit{electrons} creating \textit{electric force} fields; the \textit{Sun} and the \textit{Earth} are part of the \textit{solar system}, as the \textit{nucleus} and  \textit{electrons} are part of the \textit{atom}; \textit{Newton} discovered \textit{gravity}, as \textit{Faraday} is credited with discovering \textit{electric force}.

Note that relation is an asymmetric function, as the pairs are ordered; e.g., Newton discovered gravity, but gravity did not discover Newton. 

\begin{table}[t!]
    \begin{center}
        \begin{tabular}{ l@{\hskip 0.2in}c@{\hskip 0.4in}l }
            \hline
            \ \ \ \ \ \ \ \ \base & \textbf{Mapping} & \ \ \ \ \ \ \targ \\ 
            \hline
            Sun & $\rightarrow$ & Nucleus \\
            Earth & $\rightarrow$ & Electrons \\
            Gravity & $\rightarrow$ & Electric force \\
            Solar system & $\rightarrow$ & Atom \\
            Newton & $\rightarrow$ & Faraday \\
            \hline
        \end{tabular}
        \centering
        \caption{Illustration of a relational analogy between the solar system and the atom.}
        \label{fig:runningExample}
    \end{center}
\end{table}

Slightly abusing notation, we denote the \emph{set} of relations that hold between two entities $e_1, e_2$ as $\rel(e_1,e_2) \subseteq 2^\rel$. For example, $\rel(Earth, Sun)$ contains \{revolve around, attracted to\}, etc. 
For clarity, we {sometimes} use $\rel_B$, $\rel_T$ to emphasize that the entities belong to the \base, \targ domain.

\xhdr{Similarity}
Let \simrel be a similarity metric between two \emph{sets} of relations, $\simrel : 2^\rel \times 2^\rel \rightarrow [0, \infty)$. 
Intuitively, when applied to singletons, we want our similarity metric to capture how relations are like each other. For example, ``revolve around'' is similar to ``orbit'' and (to a lesser degree) ``spiral''. When applied to sets of relations, we want \simrel to be higher if the two sets \emph{share many distinct} relations. For example, \{revolve around, attracted to\} should be more similar to \{orbit, drawn into\} than to \{revolve around, orbit\} (as the last set does not include any relation similar to attraction). In Section \ref{subsec:singleMappingScore} we present our \simrel implementation. 

Given one pair from \base and one from \targ, we define similarity in terms of their relations. Since \rel is asymmetric, we consider both directions:
\begin{equation}
    \begin{aligned}[t]
        \simi(b_1,b_2,t_1,t_2) = \\
        \simrel(\rel_B(b_1,b_2), \rel_T(t_1,t_2)) +\\ \simrel(\rel_B(b_2,b_1), \rel_T(t_2,t_1)) \nonumber
    \end{aligned}
\end{equation}
\xhdr{Objective}
Our goal is to output a mapping $\ensuremath{\mathcal{M}}: \base \rightarrow \targ \cup \bot$ such that no two \base entities are mapped to the same \targ entity (Table \ref{fig:runningExample}). Mapping into $\bot$ means the entity was not mapped to any entity in the \targ domain. 

We look for the mapping ${\mathcal{M^*}}$ that captures the best inter-domain analogical structure similarity by maximizing the relational similarity: 
    \begin{align}
        \argmax_{\mathcal{M}} \sum_{j=1}^{n-1} & \sum_{i=j+1}^{n} \simi (b_j,b_i,\mathcal{M}(b_j),\mathcal{M}(b_i)) \nonumber
    \end{align}
\noindent Note: if $b_i$ or $b_j$ maps to $\bot$, $\simi$ is defined to be 0.

\section{Analogous Matching Algorithm}
\label{sec:analogousMatching}

We wish to find the best mapping from \base to \targ. We first extract relations between entity pairs from the same domain (Section \ref{subsec:relationExtraction}). Then, we compute the similarity between entity {pairs} that could be mapped (Section \ref{subsec:singleMappingScore}). Finally, we build the mapping (Section \ref{subsec:kernelScore}).

\subsection{Relation Extraction}
\label{subsec:relationExtraction}
Automatically extracting relations is a key part of our algorithm, as it eliminates the need for extensive manual curation of the input. We focus on \emph{commonsense} relations (e.g., the Earth \textit{revolves around} the Sun), as opposed to situational relations (e.g., the book is on the table). This broadly falls under open information extraction (OIE), the task of generating a structured representation of the information in a text. 
There has been a lot of work in this area, especially attempts to automate the construction of commonsense datasets \citep{etzioni2008open, etzioni2004web, yates2007textrunner, lenat1985cyc, sap2019atomic}.

Given two entities, 
we automatically extract relations from multiple sources: 

\xhdr{ConceptNet} A commonsense dataset, containing about $1.5$M nodes \cite{liu2004conceptnet}. For each entity, we receive a list of (predicate, entity), which we filtered to match the second entity (single or plural form). The predicates serve as our relations. 

\xhdr{Open Information Extraction} A database automatically extracted from a large web corpus \cite{etzioni2008open}. It contains over 5B triplets of the form (subject, predicate, object). We searched for a match between both entities in the (subject, object) fields, and used the predicates as our relations.

\xhdr{GPT-3 (text-davinci-001)}\footnote{GPT-3 is the only data source that is not freely available. All queries needed for this paper accumulated to less than $\$50$.} We used a generative pretrained large language model (LM) as a knowledge base in a few-shot manner \citep{petroni2019language,brown2020language}. We input a prompt of four analogies, e.g., ``Q: What are the relations between gravity and Newton?, A: Newton discovered gravity. A: Newton invented gravity.'' (see Section \ref{app:gpt} for the full prompt). GPT-3 outputs up to three sentences per query. We kept only sentences of the form <entity> <text> <entity>, treating the <text> as the relation.  

\xhdr{Quasimodo} A commonsense knowledge base that focuses on salient properties of objects \cite{romero2020inside}. It contains more than $3.5$M triplets of (subject, predicate, object). It considers \emph{questions} instead of statements. For instance, if people search for an answer to ``Why is the sky blue?'', this implies that the sky is blue. 
Whenever our two entities appeared in the (subject, object) fields, we extracted their predicates as relations.

\begin{figure}
    \includegraphics[width=0.9\linewidth]{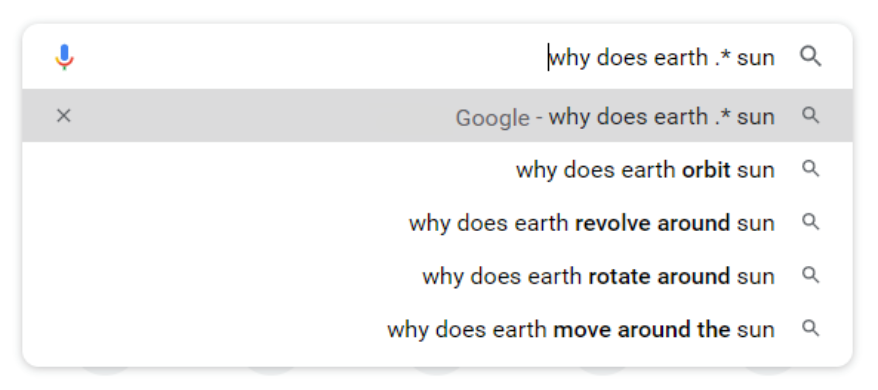}
    \caption{Quasimodo++. Example regex used to extract suggestions from Google (``<question> <entity1> .* <entity2>''). We use questions such as ``Why does'', ``Why did'' and ``How does''. }
    \label{fig:relationaGoogle}
\end{figure}

\xhdr{Quasimodo++} A relation extraction method that we develop, inspired by Quasimodo. Quasimodo was constructed using questions about a single entity; we extended it to questions exploring relations between \emph{pairs} of entities. We used Google's query auto-completion to tap into the query logs, asking questions containing both desired entities, such as ``How does Earth * Sun'', ``How is Earth * Sun'', and ``Why does Sun * Earth'' for every pair of entities (see Figure \ref{fig:relationaGoogle} for an example).
The exact regular expressions we used can be found in Section \ref{app:regex}.

\vspace{3pt}
We presented here the knowledge sources we implemented. We note that our algorithm is easy to extend to new sources and that we expect that its robustness will increase with coverage. 

\subsection{Scoring Entity Pairs}
\label{subsec:singleMappingScore}
We wish to calculate $sim^*(b_i,b_j,t_k,t_p)$ for $b_{i, j} \in \base$, $t_{k, p} \in \targ$, $1\leq i<j \leq n$, $1\leq k \neq p \leq m$.
\remove{
\begin{equation*}
    sim^*(b_i,b_j,t_k,t_p)
\end{equation*}
\begin{equation*}
    \forall \ i,j:  1 \leq i < j \leq n,\ \  \forall \ k,p: 1 \leq k \neq p \leq m
\end{equation*}
}

In Section \ref{sec:problemDefinition} we specified desiderata of $\simrel$, especially that it is higher if the two sets share many distinct relations.
We now present our implementation of $\simrel$.

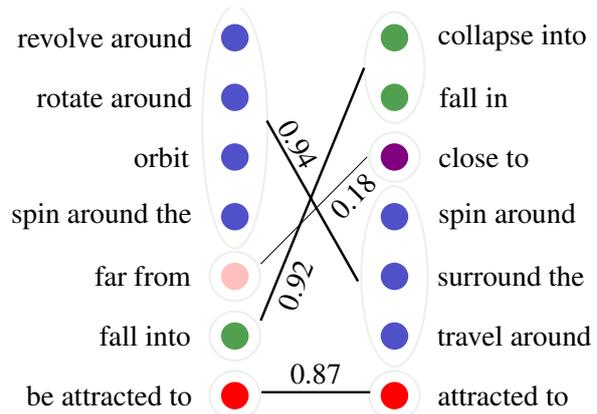
\begin{figure}[t]
    \begin{center}
        \begin{tikzpicture}[thick,
          every node/.style={circle},
          g1c1/.style={fill=blue},
          g1c2/.style={fill=pink},
          g1c3/.style={fill=green},
          g1c4/.style={fill=red},
          g2c1/.style={fill=green},
          g2c2/.style={fill=purple},
          g2c3/.style={fill=blue},
          g2c4/.style={fill=red},
        ]
            \clip (-2.95,-5) rectangle (4.7, 0.4);
        
            \begin{scope}[start chain=going below,node distance=4mm]
                \node[g1c1,on chain] (l1) [label={[label distance=0.2cm]left: revolve around}] {};
                \node[g1c1,on chain] (l2) [label={[label distance=0.2cm]left: rotate around}] {};
                \node[g1c1,on chain] (l3) [label={[label distance=0.2cm]left: orbit}] {};
                \node[g1c1,on chain] (l4) [label={[label distance=0.2cm]left: spin around the}] {};
                \node[g1c2,on chain] (l5) [label={[label distance=0.2cm]left: far from}] {};
                \node[g1c3,on chain] (l6) [label={[label distance=0.2cm]left: fall into}] {};
                \node[g1c4,on chain] (l7) [label={[label distance=0.2cm]left: be attracted to}] {};
            \end{scope}
            
            \begin{scope}[xshift=2.1cm,start chain=going below,node distance=4mm]
                \node[g2c1,on chain] (r1) [label={[label distance=0.2cm]right: collapse into}] {};
                \node[g2c1,on chain] (r2) [label={[label distance=0.2cm]right: fall in}] {};
                \node[g2c2,on chain] (r3) [label={[label distance=0.2cm]right: close to}] {};
                \node[g2c3,on chain] (r4) [label={[label distance=0.2cm]right: spin around}] {};
                \node[g2c3,on chain] (r5) [label={[label distance=0.2cm]right: surround the}] {};
                \node[g2c3,on chain] (r6) [label={[label distance=0.2cm]right: travel around}] {};
                \node[g2c4,on chain] (r7) [label={[label distance=0.2cm]right: attracted to}] {};
            \end{scope}
            
            \node [almostwhite,fit=(l1) (l4), style={ellipse,draw,inner sep=-7pt,text width=1.1cm}] {};
            \node [almostwhite,fit=(l5) (l5), style={ellipse,draw,inner sep=1pt,text width=0.4cm}] {};
            \node [almostwhite,fit=(l6) (l6), style={ellipse,draw,inner sep=1pt,text width=0.4cm}] {};
            \node [almostwhite,fit=(l7) (l7), style={ellipse,draw,inner sep=1pt,text width=0.4cm}] {};
            \node [almostwhite,fit=(r1) (r2), style={ellipse,draw,inner sep=-2pt,text width=0.7cm}] {};
            \node [almostwhite,fit=(r3) (r3), style={ellipse,draw,inner sep=1pt,text width=0.4cm}] {};
            \node [almostwhite,fit=(r4) (r6), style={ellipse,draw,inner sep=-4pt,text width=0.9cm}] {};
            \node [almostwhite,fit=(r7) (r7), style={ellipse,draw,inner sep=1pt,text width=0.4cm}] {};
            
            \draw[-,line width=0.94pt] (0.42, -1.1) -- node[sloped, above=-0.35cm, xshift=-0.75cm] {0.94} (1.62,-3.23);
            \draw[-,line width=0.18pt] (0.34, -3) -- node[sloped, below=-0.3cm, xshift=0.5cm] {0.18} (1.74,-1.58);
            \draw[-,line width=0.92pt] (0.34, -3.75) -- node[sloped, below=-0.3cm, xshift=-1.2cm] {0.92} (1.7,-0.4);
            \draw[-,line width=0.87pt] (0.35, -4.7) -- node[sloped, above=-0.3cm] {0.87} (1.78,-4.7);
        \end{tikzpicture}
        \centering
        \caption{Left: partial relations of \textit{Earth:sun}. Right: partial relations of \textit{electron:nucleus}. This is the result of the maximum weighted match on the clusters. Colors correspond to clusters.}
        \label{graph:clustering}
     \end{center}
\end{figure}

Without loss of generality, let us consider $\simrel(\rel_B(b_1,b_2), \rel_T(t_1,t_2))$.
We first extract all relations $\rel_B(b_1,b_2), \rel_T(t_1,t_2)$. Next, we calculate the score between each relation in $\rel_B(b_1,b_2)$ and each relation in $\rel_T(t_1,t_2)$. We create a complete bipartite graph where the left side nodes are the relations of $\rel_B(b_1,b_2)$, and the right side nodes are the relations of $\rel_T(t_1,t_2)$ (Figure \ref{graph:clustering}). The edge weights ($w$) are the cosine similarity of the nodes' sBERT embedding \cite{reimers2019sentence}.

We remove non-informative relations by extracting the top-frequent n-grams ($n=\{1,2,3,4\}$) from Wikipedia and setting their score to zero. Edges that did not reach a threshold (chosen using hyper-parameter search, see Section \ref{subsec:hpFt}) were set to zero.
\remove{
\begin{equation*}
    w^*(r_b,r_t) = 
    \begin{cases}
        0 &\text{$r_b \in \textit{top\_n-grams}\ \vee$}\\
         & \text{$r_t \in \textit{top\_n-grams}$} \\ 
        0 &\text{$w < \textit{threshold}$} \\
        w &\text{else}
    \end{cases}
\end{equation*}}

Next, we cluster similar relations on each side (e.g., ``revolve around'' and ``circle around'') to avoid double-counting. We use hierarchical agglomerative clustering based on the cosine embedding similarity (threshold $=0.5$; see Section \ref{subsec:hpFt}). The weight of edges between two clusters is the maximal weight of an edge between their nodes (see Figure \ref{graph:clustering}; colors correspond to clusters). 

Finally, we apply Maximum-Weight Bipartite Matching on the clusters (see Section \ref{subsec:hpFt}). The similarity score $\simrel(\rel_B(b_1,b_2), \rel_T(t_1,t_2))$ is defined as the sum of the remaining edges.

\begin{figure*}[ht!]
        \includegraphics[width=\textwidth]{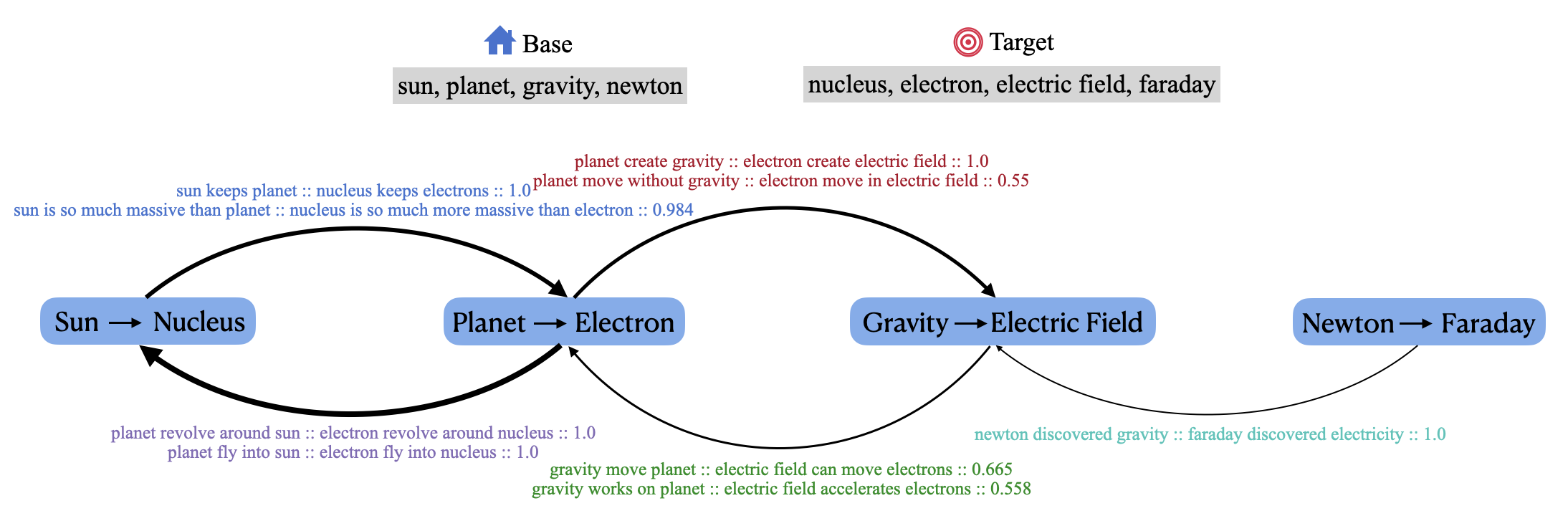}
        \caption{A snippet from our UI. Top: Input. Bottom: The mapping our algorithm found (output), is represented as a graph. Nodes correspond to mappings between single entities (e.g., sun to nucleus). Each edge is annotated with some of the shared relations between the mapped pairs corresponding to its endpoints and their similarity score. For the sake of visualization, we show at most two relations for each edge. Edge weight corresponds to strength.}
            \label{graph:full_mapping}
\end{figure*}

\subsection{Building a Mapping}
\label{subsec:kernelScore}
Using the score mappings between pairs, we can compose larger mappings.
We use beam-search, starting from the most promising pair-mappings found in Section \ref{subsec:singleMappingScore}. In each iteration, we expand the $20$ most promising partial mappings, testing each possible mapping between single entities of \base and \targ (that are consistent with the current partial mapping -- i.e., a \base entity cannot map to multiple \targ entities). When expansions stop increasing the score, we stop the search and select the mapping with the highest score.

Figure \ref{graph:full_mapping} shows a snippet from our UI. Input appears on the top. {\fame}'s output mapping is represented as a graph: nodes correspond to single entity mappings (e.g., Sun to nucleus). Edges represent the shared relational structure. 
Each edge contains some of the shared relations between the mapped pairs corresponding to its endpoints (e.g., ``more massive than'') and their similarity score (note the edges are directional). To ease visualization, we show at most two relations per edge. The thickness of an edge corresponds to its weight.

\remove{Kernel is a structure of one or more single-mappings that do not contradict each other\footnote{Example contradictory structure: \{$b_1$ $\rightarrow$ $t_1$, $b_2$ $\rightarrow$ $t_2$, $b_1$ $\rightarrow$ $t_2$, $b_3$ $\rightarrow$ $t_3$\}.\label{fn:kernelCont}}. Each kernel receives a structure score: 
\begin{equation}
    Kernel\ Score = \sum_{j=1}^{n-1} \sum_{i=j+1}^{n} sim^*(b_j,b_i,t_j,t_i)
\end{equation}

We now turn to score a kernel. Let K be the current kernel:
\begin{equation*}
    \base = \{b_1, ..., b_n\},  \targ = \{t_1, ..., t_n\}
\end{equation*}
\begin{equation*}
    s.t \ \forall i \in [n],  b_i \rightarrow t_i
\end{equation*}
We denote the previous kernel score as $s_n$ and calculate the kernel score in a recursive manner (in the base case of $n=2$, $s_n$ is the single mapping score).  Given a new pair of entities from \base and \targ that map each other $b_{n+1} \rightarrow t_{n+1}$, we calculate all single-mappings derived from $b_{n+1}$ and $t_{n+1}$:
\begin{equation*}
    x = \sum_{i=1}^{n} sim^*(b_i,b_{n+1},t_i,t_{n+1})
\end{equation*}
The kernel score is $S_{n+1} = S_n + x$. We note that this recursive calculation is done to include the relations of the $n+1$ entity with the relations of all $[n]$ entities in the structure. Thus, we not only add the new single mapping score, but perform an holistic calculation of the entire structure's strength. \cnote{The motivation behind the recursive calculation is terrible, rewrite}
}

\xhdr{A note on the solution space} 

In previous works $n=m$ and \ensuremath{\mathcal{M}} is a \textit{bijective} function. Meaning, \ensuremath{\mathcal{M}} is both injective (one-to-one; each element in the target is the image of at most one element in the source) and surjective (onto; all the target terms are covered). In other words, no entity is left unmapped. In that case, the solution space's cardinality is $n!$.

We allow for $n \neq m$ and for entities to remain unmapped. Without loss of generality let $n \leq m$.
The cardinality is then $\left(\sum_{i=0}^{n} \binom{n}{i} \frac{m!}{(m-i)!}\right)-(n\cdot m)$, where ${i}$ is the number of matched entities. 
We subtract $n\cdot m$ because we do not allow for a mapping of size 1; our algorithm starts by mapping pairs and then adds single-entity mapping at each iteration of the beam search. 

This relaxation of the bijective constraint drastically increases the space; for $n=7$, $n! = 5,040$, while our space is of size $130,922$. 

\label{sec:complexity}

\xhdr{Hyper-Parameter Search}
\label{subsec:hpFt}
We constructed a new dataset to set our model's hyper-parameters (See Appendix \ref{app:parameters}). The dataset contains $36$ analogical mapping problems created by ten volunteers, not from our research team. We showed them example analogies and asked them to generate new ones. An expert from our team verified their output, discarding $4$ analogies. Domain size varied between 3 to 5 (average size=$3.4$).

On the problems generated by the volunteers, \fame \ achieves $83.3\%$ perfect mappings (the whole mapping is correct). If we consider single mappings separately, it achieves $89.4\%$ accuracy.


\section{Entity Suggestion}
\label{sec:entitiesSuggestions}
One of the main limitations of previous analogical mapping algorithms is their inability to automatically expand analogies.
This is especially interesting in our case, as we allow for unmapped entities; thus, suggesting new entities could identify potential mapping candidates for the unmapped entities.  

For example, let \base = \{Sun, Earth, gravity, Newton\} and \targ = \{nucleus, electron, electricity\}. The correct mapping is Sun $\rightarrow$ nucleus, Earth $\rightarrow$ electron, {gravity} $\rightarrow$ electricity, leaving Newton with no mapping. Our goal is to suggest candidate entities that preserve the relational structure. 

Intuitively, we look at the relations Newton shares with other \base entities (e.g., discovered gravity), and try to see which \targ entity plays a corresponding role (i.e., who discovered electricity?). 

\remove{
\begin{figure}[ht]
    \begin{center}
        \begin{tabular}{ l@{\hskip 0.5in}l }
            \hline
            \textbf{Base} & \textbf{Target} \\ 
            \hline
            gravity & electrons \\
            sun & electricity \\
            newton & nucleus \\
            Earth &  \\
            \hline
        \end{tabular}
        \centering
        \caption{Example of \base and \targ with different size, our system can deal with it}
        \label{table:input-suggstions}
    \end{center}
\end{figure}
}

\remove{
\begin{figure}[ht]
    \begin{center}
        \begin{tabular}{ l@{\hskip 0.2in}c@{\hskip 0.4in}l }
            \hline
            \textbf{Base} & \textbf{Mapping} & \textbf{Target} \\ 
            \hline
            sun & $\rightarrow$ & nucleus \\
            Earth & $\rightarrow$ & electrons \\
            gravity & $\rightarrow$ & electricity \\
            newton & $\rightarrow$ & ? \\
            \hline
        \end{tabular}
        \centering
        \caption{The system identifies the existing mapping, and now it will search for suggestions. In that case, the suggestions are: \textit{Benjamin Franklin}, \textit{Nikola Tesla}, \textit{Michael Faraday}}
        \label{table:output-suggstions}
    \end{center}
\end{figure}  
}

More formally, suppose we wish to find candidates $t^*$ for mapping to $b_{n}$.
We first extract the relations of $R_b(b_i,b_{n})$, $\forall i \in [n]$ (denoted as $R_{b_i}$).
We then iterate over all relations $r \in R_{b_i}$ and use the pair $\{\mathcal{M}(b_{i}), r\}$ to extract suggestions for $t^*$. 

We use Open Information Extraction, Quasimodo, and Quasimodo++. While our method was previously used to extract \textit{relations} given a pair of two entities, we now use it to extract \textit{entities} given a pair of \{entity, relation\}. This entails filtering on the predicate field in our commonsense datasets and changing the queries in Quasimodo++.

\remove{
\dnote{get comments, in particular input/output in all your pseudocode. meaningful names -- what's up with single-letter variables?}
\begin{algorithm}
	\caption{Suggestions extraction} 
	\begin{algorithmic}[1]
	    \State $suggestions \leftarrow []$
		\For {$i \in [n]$} \Comment{n is the size of the current kernel}
			\For {$r \in  R_b(b_i, b_{n+1})$}
    			\State $suggestions \leftarrow suggestions \cup s(t_i,r)$ \Comment{Notice that t_i }
    		\EndFor
		\EndFor
		\State return $suggestions$
	\end{algorithmic} 
	\label{algo:suggestionExtraction}
\end{algorithm}
}

As suggestions tend to be noisy, we cluster all extracted entities (similarly to the clustering from Section \ref{subsec:singleMappingScore}). We remove clusters of size $<2$.

For each suggestion cluster, we rerun our analogous matching algorithm with a representative entity from that cluster (the closest to the cluster's center of mass). We pick the cluster whose representative resulted in the mapping with the highest score.
As the commonsense datasets we work with operate mostly on string matching, small changes (e.g., Benjamin Franklin/Ben Franklin) could sometimes result in slightly different results. Thus, we perform one final round, with \textit{all} entities from our chosen cluster, and pick the highest score mapping.

\section{Evaluation}
In this section, we evaluate {\fame}. We test its ability to identify the correct mapping (Section \ref{sec:performance}), and compared it to both related works (Section \ref{sec:compSme}) and human performance (Section \ref{sec:compPeople}).

\subsection{Performance on Analogy Problems}
\label{sec:performance}

\begin{table}[t]
    \begin{tabular}{p{2.83cm}|p{0.9cm}|p{0.9cm}|p{0.9cm}}
        \textbf{Sources} & \textbf{Near} & \textbf{Far} &\textbf{Extended} \\
        \hline
         All & $\mathbf{85\%}$ & $\mathbf{77.5\%}$ & $\mathbf{77.8\%}$ \\ \hline 
         All-ConceptNet & $\mathbf{85\%}$ & $\mathbf{77.5\%}$ & $\mathbf{77.8\%}$  \\
         All-Open IE & $\mathbf{85\%}$ & $67.5\%$ & $58.3\%$ \\
         All-Quasimodo & $\mathbf{85\%}$ & $\mathbf{77.5\%}$ & $72.2\%$ \\
         All-Quasimodo++ & $80\%$ & $72.5\%$ & $72.2\%$ \\
         All-GPT-3 & $57.5\%$ & $50\%$ & $66.7\%$ \\
    \end{tabular}
    \caption{Ablation study on the 2x2 near and far problems and our extended set, leaving out knowledge sources. Results show the {importance of the generative LM approach} (GPT-3.5) as a knowledge source. {Open Information Extraction also contributes much}, especially for the complex analogies (2x2-far and extended). 
    }
    \label{tab:ablation}
\end{table}

\xhdr{2x2 problems}
One of the things that might have held computational analogy back is the lack of high-quality, large-scale datasets. Most datasets are small and focus on classical 2x2 problems ($A:B::C:D$), similar to SAT questions. 

We start by testing {\fame} on this standard type of analogies. 
We use $80$ problems from \citet{green2010connecting}, split into $40$ near and $40$ far analogies (e.g., for ``answer:riddle'', near analogy is ``solution:problem'', far analogy is ``key:lock''). 
While the dataset is small, we believe it is still interesting to explore. 
Our algorithm managed to perfectly map $85\%$ of near analogies and $77.5\%$ of far ones. Random guess baseline is $33.3\%$ (see Section \ref{subsec:kernelScore}).

\xhdr{Extended problems}
Encouraged by the results of the 2x2 problems, we explore more complex problems. We decided to extend the Green far analogies (which are harder than the near ones). We had three experts go over the dataset together and brainstorm potential extensions. On four problems, the experts did not manage to agree on any additional mappings, leaving us with $36$ extended problems (average domain size $3.3$).  

Our algorithm perfectly mapped  $77.8\%$ of the extended problems. Random baseline is $13.1\%$ on average. As we relax the bijection assumption, {\fame}'s average guess level is even lower -- $2.2\%$ (see Section \ref{sec:complexity}). If we look beyond the top-rated solution, our algorithm has the correct solution in its top-2 guesses $83.3\%$ of the time and $91.7\%$ for top-3. 

\xhdr{Error analysis}
We found 3 main causes of error: 
\begin{compactitem}
\item \textbf{Coverage} (for example, we could not find a relation between ``hoof'' and ``hoofprint'').
This prompted us to ablate the knowledge sources {\fame} uses (Table \ref{tab:ablation}). Results show the importance of the generative LM approach. Open IE is also important, especially for the more complex analogies (far and extended). Some sources, such as ConceptNet, did not seem to contribute much. 
\item \textbf{Noisy relations} that are either peculiar or plain wrong (e.g., ``a footballer can iron'').
\item \textbf{Embedding similarity} (for example, ``produce'' and ``is produced by'' have a high similarity score). 
This is exacerbated by {\bf ambiguity} (e.g., the word ``pen'' referred to ``pigpen'' and not to the writing instrument).
\end{compactitem}

\subsection{Comparison to Related Work}
\label{sec:compSme}

\xhdr{SME line of work} We had difficulty comparing {\fame} to SME \citep{falkenhainer1989structure} and its extensions, due to their complex input requirements. 
LRME \citep{turney2008latent} is closest to our setting, but no code or demo is available. Thus, we compare to their published results on a set of $20$ problems. 

LRME's entities include nouns, verbs, and adjectives. Since {\fame} expects noun phrases, we filtered out all other input terms (one problem has only a single noun, so we are left with $19$ problems). It is hard to compare in this setup (and unfortunately, authors did not report which partial mappings were correct). Still, LRME's accuracy was $75\%$, whereas \fame \ achieved $84.2\%$. 

While the size of the problems is smaller when restricted to nouns, we believe the noun-only setting is harder. The verbs and adjectives often provide hints that significantly constrain the search space. For example, in problem A6 \citep{turney2008latent}  (mapping a projectile to a planet) there is one adjective in each domain (parabolic, elliptical). Those adjectives can only apply to one or two of the nouns (i.e., you cannot have parabolic earth, air, or gravity), effectively giving away the noun mapping. 

As a side note, we also believe that our noun-only input is a cleaner problem setting, as it is often easier to automatically identify the entities in a domain than to identify the attributes and verbs relevant for the analogy. In the words of LRME's authors,  ``LRME is not immune to the criticism of \citet{chalmers1992high}, that the human who generates the input is doing more work than the computer that makes the mapping.'' We believe  {\fame} is a step in the right direction in this regard.

\xhdr{Pretrained LMs}
In the absence of a baseline, we turn to a generative pretrained large LM known to have impressive commonsense abilities -- GPT-3.5 (text-davinci-002). We used 4 random examples from the hyper-parameter search dataset. After some experimentation with prompt engineering, we chose two variants (see \ref{app:gpt}).

The results are summarized in Table \ref{tab:baseline}. GPT-3.5 does well on the 2x2 datasets \citep{green2010connecting}. However, both datasets appear on the web, and perhaps GPT-3.5 was exposed to them during training (data leakage). In particular, we found some of the answers via a simple web search (Figure A.\ref{fig:noseAnalogy}). 

Moreover, GPT-3.5's performance drops on the extended set, where problems are complex and do not appear on the web. Interestingly, it does not even manage to return a valid mapping in some of the cases. This exercise improves our understanding of {\fame}'s strengths and weaknesses. 

\begin{table}[t]
    \begin{tabular}{l|c|c|c}
        \textbf{Algorithm} & \textbf{Near} & \textbf{Far} & \textbf{Extended} \\
        \hline
         \fame & ${85.0\%}$ & ${77.5\%}$ & $\mathbf{77.8\%}$ \\ 
         GPT-3.5 ``:'' & $\mathbf{92\%}$ & $\mathbf{80\%}$ & $44\%$ \\
         GPT-3.5 ``-->'' & $88\%$ & $\mathbf{80\%}$ & $58\%$ \\ 
    \end{tabular}
    \caption{Comparison of {\fame} and GPT-3.5. GPT-3.5 does well on the 2x2 datasets (far and near). We note that data leakage is a concern. GPT-3.5's performance sharply drops on the extended problem set, where problems are bigger and do not appear on the web.}
    \label{tab:baseline}
\end{table}

\remove{
\begin{table}[ht]
    \begin{center}
        \begin{tabular}{l@{\hskip0.2in}c@{\hskip0.4in}l@{\hskip 0.2in}l}
            \hline
            \ \ \ \base & \textbf{Mapping} & \ \ \ \ \targ & POS \\
            \hline
            Balls & $\rightarrow$ & Bullets & NN(P) \\
            Bat & $\rightarrow$ & Gun & NN \\
            Round & $\rightarrow$ & Aerodynamic & ADJ \\
            \hline
        \end{tabular}
        \centering
        \caption{Made up illustration of hints provided by input using the setup described in LRME \citep{turney2008latent}. These caveats motivated us to use only singular nouns as input. In this example, it is possible to use the plural hint to map Balls to Bullets. Moreover, algorithms supporting adjectives in their input can harness the fact that Balls are round (unlike Bat) and Bullets are aerodynamic (unlike Gun). This results in a mapping between Balls to Bullets and Round to Aerodynamic. Using the bijection assumption all previous works relied on (and we relax here), Bat must be mapped to Gun.}
        \label{fig:lrmeFail}
    \end{center}
\end{table}  
}

\xhdr{E-KAR dataset} 
\citet{chen2022kar} recently released a relevant dataset, E-KAR, for rationalizing analogical reasoning. The dataset consists of multiple-choice problems from civil service exams in China. For example, for the source triplet ``tea:teapot:teacup'', the correct answer is  ``talents:school:enterprise''. The reasoning is that both teapot and teacup are containers for tea. After the tea is brewed in the
teapot, it is transported into the teacup. Similarly, both school and enterprise are organizations. After talents are educated in school, they are transported into enterprise.\footnote{Interestingly, the authors of this paper thought that the ``passengers:bus:taxi'' answer was the correct one, based on containment and size relations.}

The E-KAR test set has no labels, so we used their validation set (N=119) to test \fame. As our task is different, we only took source entities (as \base) and entities from the correct answer (as \targ). 
We filtered questions without nouns, resulting in N=101. 

{\fame} found the right mapping $68.3\%$ of the time. A closer examination of {\fame}'s mistakes revealed that $\sim 75\%$ of them occurred due to relation \emph{types} that are not at all covered by our framework: either ternary relations (soldier:doctor:military doctor $\rightarrow$ car:electric vehicle:electric car; the last term is a combination of the first two) or relations based on sharing some attribute (so ``both containers for holding tea'' is mapped to ``both are organizations''). Some of the attribute-based mappings work at the whole-set level, so each entity on \base could map to each entity on \targ (yellow:red:white $\rightarrow$ sad:happy:angry). Thus, we conclude there is a big gap between {\fame} and E-KAR's assumptions.
\label{sec:evaluation}

\subsection{Comparison to People}
\label{sec:compPeople}

We compare {\fame} with \emph{human thinking} in a 2-phase experiment.\footnote{The experiment received ethics committee approval. See full instructions in Section \ref{app:experiments}.} In the closed-world phase, the participants received ten structure mapping problems, in which they were asked to match instances from \base to \targ. The domains included between $3$-$5$ entities (Table A.\ref{fig:part1problems}).
Participants were instructed to map each \base entity into exactly one \targ entity. 
 
In the open-world phase, participants received five \textit{mapped} problems, but one entity was left blank (Table A.\ref{fig:part2example}). Participants were instructed to fill in the blank with an entity that preserves the analogy. 

\xhdr{Participants}
We recruited $304$ participants using social media. The compensation was a chance to win one of three $\$30$  vouchers. $76.6\%$ of our participants were between the ages $18$-$35$ and $17.2\%$ are between $36$-$45$ (self-reported).

\xhdr{Closed-world mapping}
\fame \ missclassified one problem compared to gold standard ($A9$, Table A.\ref{fig:part1problems}), achieving $90\%$ accuracy (human baseline was $70.2\%$; see full distribution in Table A.\ref{fig:part1problems}).

Problem $A6$ has the lowest human accuracy ($35.5\%$), and is also the largest one ($|\base|=|\targ|=5$). A closer examination of its confusion matrix reveals that while {\fame} correctly mapped \textit{water} to \textit{heat} and \textit{pressure} to \textit{temperature}, $15\%$ of people switched the two. This might be due to the strong semantic pairing of \textit{water} and \textit{temperature}. {\fame} is immune to this, as it relays on \emph{relations}.

On average, each participant mapped the problem the same as \fame \ $78\%$ of the times. 
Overall, {\fame} outperforms humans, and most of the disagreement is due to human errors.

\xhdr{Open-world entity suggestion}
We presented participants with five \textit{mapped} problems where one entity was left blank (Table A.\ref{fig:part2example}) and asked them to fill in the black while preserving the analogy.

For all five problems, an entity from {\fame}'s top two completions appeared in humans' top three completions (Table A.\ref{fig:part2problems}). Meaning, our algorithm's top suggestions are similar to humans'. 
Only in one example ($B5$) one of the top two algorithm's completions appeared \textit{third} in humans' (in the rest it is first or second). We suspect that this confusion in $B5$ occurred because \textit{gravity} and \textit{Newton} reminded participants of the term \textit{apple}. 

\begin{figure}
    \centering
    \includegraphics[width=\linewidth]{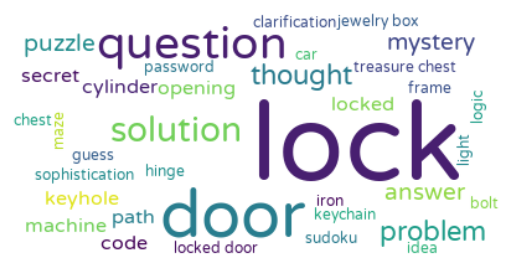}
    \caption{Word cloud of human completions for $B1$ (Table A.\ref{fig:part2problems}). While most responses were from the same semantic domain, some were creative and appropriate (e.g., treasure chest, jewelry box, car).}
    \label{fig:wordCloud}
\end{figure}

Figure \ref{fig:wordCloud} shows a word cloud for answers to problem $B1$. While most responses are quite similar, some participants returned creative and appropriate solutions (e.g., treasure chest, jewelry box, car).

\remove{

\section{End-to-end Example}
Let B = \{sun, planet, gravity, newton\} and T = \{nucleus, electrons, electricity, faraday\}. We will describe the high-level flow of the process. At each step (from step 2), we will illustrate the current state of the best solution that future to come.

\vspace{0.3cm}
\textbf{Step 1} \\
Calculate all the single mappings. Assign a vector for them called V. Let us take a look on a single direction in $sun$:$planet \sim nucleus$:$electrons$. After filtering the non-interesting relations (such as \textit{related to} or \textit{affect}), we getting the following graph:
\begin{figure}[]
    \begin{center}
        \begin{tikzpicture}[thick,
          every node/.style={circle},
          g1c1/.style={fill=green},
          g1c2/.style={fill=blue},
          g1c3/.style={fill=purple},
          g2c1/.style={fill=blue},
          g2c2/.style={fill=green},
          g2c3/.style={fill=purple},
        ]
            \clip (-2.7,-9.2) rectangle (7.5, 0.4);
        
            \begin{scope}[start chain=going below,node distance=\topbuttonpaddingnodes]
                \node[g1c1,on chain] (l1) [label={[label distance=\textpaddingnodes]left: be closer to}] {};
                \node[g1c1,on chain] (l2) [label={[label distance=\textpaddingnodes]left: is between}] {};
                \node[g1c1,on chain] (l3) [label={[label distance=\textpaddingnodes]left: is closest to}] {};
                \node[g1c1,on chain] (l4) [label={[label distance=\textpaddingnodes]left: is near}] {};
                
                \node[g1c2,on chain] (l5) [label={[label distance=\textpaddingnodes]left: crash into}] {};
                \node[g1c2,on chain] (l6) [label={[label distance=\textpaddingnodes]left: fly into}] {};
                \node[g1c2,on chain] (l7) [label={[label distance=\textpaddingnodes]left: hit}] {};
                
                \node[g1c3,on chain] (l8) [label={[label distance=\textpaddingnodes]left: go around}] {};
                \node[g1c3,on chain] (l9) [label={[label distance=\textpaddingnodes]left: move around}] {};
                \node[g1c3,on chain] (l10) [label={[label distance=\textpaddingnodes]left: revolve around}] {};
                \node[g1c3,on chain] (l11) [label={[label distance=\textpaddingnodes]left: rotate around}] {};
                \node[g1c3,on chain] (l12) [label={[label distance=\textpaddingnodes]left: rotates around}] {};
                \node[g1c3,on chain] (l13) [label={[label distance=\textpaddingnodes]left: travel around}] {};
            \end{scope}
            
            \begin{scope}[xshift=3cm,start chain=going below,node distance=\topbuttonpaddingnodes]
                \node[g2c1,on chain] (r1) [label={[label distance=\textpaddingnodes]right: fall in}] {};
                \node[g2c1,on chain] (r2) [label={[label distance=\textpaddingnodes]right: fall into}] {};
                \node[g2c1,on chain] (r3) [label={[label distance=\textpaddingnodes]right: fly into}] {};
                \node[g2c1,on chain] (r4) [label={[label distance=\textpaddingnodes]right: go into}] {};
                
                \node[g2c2,on chain] (r5) [label={[label distance=\textpaddingnodes]right: is bound to}] {};
                \node[g2c2,on chain] (r6) [label={[label distance=\textpaddingnodes]right: is close to}] {};
                
                \node[g2c3,on chain] (r7) [label={[label distance=\textpaddingnodes]right: arranged around}] {};
                \node[g2c3,on chain] (r8) [label={[label distance=\textpaddingnodes]right: are orbiting}] {};
                \node[g2c3,on chain] (r9) [label={[label distance=\textpaddingnodes]right: orbiting around}] {};
                \node[g2c3,on chain] (r10) [label={[label distance=\textpaddingnodes]right: move around}] {};
                \node[g2c3,on chain] (r11) [label={[label distance=\textpaddingnodes]right: revolve around}] {};
                \node[g2c3,on chain] (r12) [label={[label distance=\textpaddingnodes]right: rotate around}] {};
                \node[g2c3,on chain] (r13) [label={[label distance=\textpaddingnodes]right: stay around}] {};
                \node[g2c3,on chain] (r14) [label={[label distance=\textpaddingnodes]right: travel around}] {};
            \end{scope}
            
            
            \node [almostwhite,fit=(l1) (l4), style={ellipse,draw,inner sep=-7pt,text width=1cm}] {};
            \node [almostwhite,fit=(l5) (l7), style={ellipse,draw,inner sep=-5pt,text width=0.8cm}] {};
            \node [almostwhite,fit=(l8) (l13), style={ellipse,draw,inner sep=-11pt,text width=1.4cm}] {};
            
            \node [almostwhite,fit=(r1) (r4), style={ellipse,draw,inner sep=-7pt,text width=1cm}] {};
            \node [almostwhite,fit=(r5) (r6), style={ellipse,draw,inner sep=-3pt,text width=0.6cm}] {};
            \node [almostwhite,fit=(r7) (r14), style={ellipse,draw,inner sep=-15pt,text width=1.7cm}] {};
            
            \draw[-,line width=0.771pt] (0.42, -0.85) -- node[sloped, above=-0.35cm, xshift=0.7cm] {0.771} (2.68,-2.55);
            \draw[-,line width=1pt] (0.34, -2.7) -- node[sloped, above=-0.2cm, xshift=-0.7cm] {1.0} (2.64,-0.9);
            \draw[-,line width=1pt] (0.5, -5.25) -- node[sloped, above=-0.2cm] {1.0} (2.43,-5.25);
        \end{tikzpicture}
        \centering
        \caption{In the left: relations of sun:planet, in the right: relations of nucleus:electrons. This graph show the result of the maximum weighted match on the clusters. Notice that in this example we taking the top-2 edges, so the score of this graph is 2, which is the maximum score. In addition, this only one direction of sun:planet and nucleus:electrons, so in practice we need to sum it with the other direction.}
        \label{graph:end-to-end-single-mapping}
    \end{center}
\end{figure}

\textbf{Step 2} \\
Extract the top N single mappings from V. For each one of them, open a new solution and:
\begin{itemize}
    \item[-] Add the corresponding entities to that solution.
    \item[-] Update $V_i$ (where V is a \textit{deepcopy} for each solution, call it from now $V_i$)
\end{itemize}

\begin{center}
    \begin{tabular}{ l@{\hskip -0.5cm}c@{\hskip -0.7cm}l }
        \hline
        & \textbf{Single solutions} & \\ 
        \hline
        \textbf{sun:planet} & $\sim$ & \textbf{nucleus:electrons} \\
        \hline \\
        \hline
        & \textbf{Current mapping} & \\ 
        \hline
        \textbf{sun} & $\rightarrow$ & \textbf{nucleus} \\
        \textbf{planet} & $\rightarrow$ & \textbf{electrons} \\
        \hline
    \end{tabular}
\end{center}

\vspace{0.5cm}
\textbf{Step 3} \\
For each one of the N solution from previous step:
\begin{itemize}
    \item[-] Extract the top N single mappings from $V_i$.
    \item[-] Add the corresponding entities to that solution.
    \item[-] update $V_i$
\end{itemize}

\begin{center}
    \begin{tabular}{ l@{\hskip -0.5cm}c@{\hskip -0.7cm}l }
        \hline
        & \textbf{Single solutions} & \\ 
        \hline
        sun:planet & $\sim$ & nucleus:electrons \\
        \textbf{planet:gravity} & $\sim$ & \textbf{electrons:electricity} \\
        \hline \\
        \hline
        & \textbf{Current mapping} & \\ 
        \hline
        sun & $\rightarrow$ & nucleus \\
        planet & $\rightarrow$ & electrons \\
        \textbf{gravity} & $\rightarrow$ & \textbf{electricity} \\
        \hline
    \end{tabular}
\end{center}

\vspace{0.5cm}
\textbf{Step 4} \\
For each one of the N solution from previous step:
\begin{itemize}
    \item[-] Extract the top N single mappings from $V_i$.
    \item[-] Add the corresponding entities to that solution.
    \item[-] update $V_i$
\end{itemize}

\begin{center}
    \begin{tabular}{ l@{\hskip -0.5cm}c@{\hskip -0.7cm}l }
        \hline
        & \textbf{Single solutions} & \\ 
        \hline
        sun:planet & $\sim$ & nucleus:electrons \\
        planet:gravity & $\sim$ & electrons:electricity \\
        \textbf{gravity:newton} & $\sim$ & \textbf{electricity:faraday} \\
        \hline \\
        \hline
        & \textbf{Current mapping} & \\ 
        \hline
        sun & $\rightarrow$ & nucleus \\
        planet & $\rightarrow$ & electrons \\
        gravity & $\rightarrow$ & electricity \\
        \textbf{newton} & $\rightarrow$ & \textbf{faraday} \\
        \hline
    \end{tabular}
\end{center}

\vspace{0.5cm}
\textbf{Step 5} \\
Extract the best solution from the N solutions using the full-mapping score calculation. The full results can be shown in Figure 7 \cnote{ref}.
\begin{figure*}[ht]
    \begin{center}
        \includegraphics[width=\textwidth]{images/full_mapping_keynote_domains.png}
        \caption{Example of the output graph. Each edge represent a direction in the single-mapping. Each line in the edges represent a cluster. There are at most three clusters for each direction. The score of the cluster also specify from the right.}
            \label{graph:full_mapping}
    \end{center}
\end{figure*}
\label{sec:fullExample}
}

\section{Related Work}
Computational analogy-making dates back to the 1960s \citep{evans1964heuristic, reitman1965cognition}. 
Analogy-making approaches are broadly categorized as symbolic, connectionist, and hybrid \citep{french2002computational, mitchell2021abstraction, gentner2011computational}.

Symbolic approaches usually represent input as structured sets of logic statements. Our work falls under this branch, as well as SME \citep{falkenhainer1989structure} and its follow-up work. LRME \citep{turney2008latent} is the closest to our work, as it automatically extracts the relations. Unlike {\fame}, LRME requires exact matches of relations across different domains. We also focus on nouns only, making the problem harder, and relax the bijection assumption, allowing for automatically extending analogies.

\xhdr{NLP} Analogy-making received relatively little attention in NLP. 
The best-known task is word analogies, often used to measure embeddings' quality (inspired by Word2Vec's {\it ``king - man + woman = queen''} example \citep{mikolov2013distributed}).
Follow-up work explored embeddings' linear algebraic structure \citep{arora2016latent, gittens2017skip, pmlr-v97-allen19a} or compositional nature \cite{chiang-etal-2020-understanding}, neglecting relational similarity.   
A recent work on analogies between procedural texts \cite{sultan2022life} did study relational similarity, but extracted the relations from the input texts, with no commonsense augmentations.

Recently, there have been efforts to study LMs' analogical capabilities \cite{ushio2021bert,NEURIPS2020_1457c0d6}. Findings indicate they struggle with
abstract and complex relations and results depend strongly on LM's architecture and parameters. 

\citet{kittur2019scaling} combined NLP and crowds for product analogies without explicitly modeling entities and relations, but instead automatically extracting \emph{schemas} of the product. 

\label{sec:relatedWork}

\section{Conclusions and Future Work}
Detecting deep structural similarity across distant domains and transferring ideas between them is central to human thinking.
We presented \fame, a novel method for analogy making. 
Compared to previous works, \fame \ is more expressive, scalable, robust and interpretable. It also allows partial matches and automatic entity suggestions to extend the analogies. 

\fame \ correctly maps $81.2\%$ of classical 2x2 analogy problems. On larger problems, it achieves $77.8\%$ perfect mappings (mean guess level=$13.1\%$). \fame \ also outperforms humans in solving analogy mapping problems ($90\%$ vs. $70.2\%$). Interestingly, our automatic suggestions of new entities resemble those suggested by humans.

In future work, we plan to improve coverage and extend our framework to more than just binary relations, as sometimes the key to an analogy is a relation involving more than two objects.
In addition, we plan to improve our similarity measure, to address both context (to solve ambiguity) and the difference between active and passive relations. We plan to explore different forms of input, such as algorithms that take as input very partial domains, perhaps even just domain \emph{names} (e.g., solar system, atom) and populate the domains with entities, or algorithms incorporating \emph{user feedback}.

To conclude, we hope \fame \ will pave the way for analogy-making algorithms that require less-restrictive inputs and can scale up and tap into the vast amount of potential inspiration the web offers, augmenting human creativity.

\label{sec:conclusions}


\section{Ethical Considerations \& Limitations}
While \fame \ can assist humans by inspiring non-trivial solutions to problems, it has been shown that humans struggle with detecting caveats in presented analogies \citep{holyoak1995mental}. For example, the cardiovascular system is often taught to medical students in terms of water supply system \citep{swain2000water}. However, this analogy might also confuse them, as it ignores important differences between water and blood (e.g., blood clots). 
Thus, while our output is interpretable, it might still mislead people, and it is important to alert the users to this possibility. 

Another issue is the fact that {\fame}'s coverage highly depends on external resources (ConceptNet, Google AutoComplete, etc.). This might be particularly problematic when applied to low-resource languages. As the relations we look for are commonsense relations, rather than cultural or situational ones, using automatic translation might ameliorate the problem. 

Lastly, we also note these resources evolve over time, and thus if one is interested in reproducibility, it is necessary to save snapshots of the extracted relations.

\label{sec:ethics}

\section*{Acknowledgements}
We thank the reviewers for their insightful comments. This work was supported by the European Research Council (ERC) under the European Union’s Horizon 2020 research and innovation programme (grant no. 852686, SIAM).

\vspace{10pt}

\begin{tcolorbox}[top=1.5pt,bottom=1pt,titlerule=0pt,colback=grey!15!white,colframe=black,arc=0pt,outer arc=0pt]
\begin{center}
\vspace{5pt}
In memory of the more than one thousand victims of the horrific massacre carried out by Hamas terrorists on October 7th, 2023.
\vspace{5pt}
\end{center}
\end{tcolorbox}

\bibliography{anthology,custom}
\bibliographystyle{acl_natbib}

\clearpage

\appendix
\label{appendix}

\section{Implementation Details}
We fine-tune our model using 36 problems described in Section \ref{subsec:hpFt}.

We used the pre-trained model \textit{msmarco-distilbert-base-v4} which is based on sBERT \cite{reimers2019sentence}. We set the similarity threshold (the similarity between two relations) to be \textbf{0.2} (range checked: 0-0.6). We set the number of top n-grams which was filtered (the top frequencies n-grams in Wikipedia) to \textbf{500}. The clustering distance threshold is set to \textbf{0.5} (range checked: 0.3-0.9). The number of clusters we consider when computing the sum is set to \textbf{3} (range checked: 1-maximum number of clusters). We set the beam search size to \textbf{20} (range checked: 1-40). All of these parameters describes in Section \ref{sec:analogousMatching}. 

We provide access to our anonymous repository can be found\footref{ft:repoLink}. We note that the usage of Docker is not supported in this version for the purpose of maintaining anonymity. However, the algorithmic content is available.
\label{app:parameters}

\subsection{Quasimodo++ regular expressions}
We use the following regex for our Quasimodo++: ``<question> <prefix> <entity1> .* <entity2>''. The questions we used are: \{``why do'', ``why is'', ``why does'', ``why does it'', ``why did'', ``how do'', ``how is'', ``how does'', ``how does it'', ``how did''\}. The prefix is optional and can be \{``a'', ``an'' and ``the''\}. We use both singular and plural forms of the entities.
\label{app:regex}

\subsection{GPT-3}
\subsubsection{Prompts used for relation extraction}

The prompt used for GPT-3 is:

\noindent Q: What are the relations between a blizzard and snowflake? \\
A: A blizzard produces snowflakes. \\
A: A blizzard contains a lot of snowflakes. \hfill \break

\noindent Q: What are the relations between an umbrella and rain? \\
A: An umbrella protects from rain. \\
A: An umbrella provides adequate protection from rain. \hfill \break

\noindent Q: What are the relations between a movie and screen? \\
A: A movie displayed on a screen. \\
A: A movie can be shown on a screen. \hfill \break

\noindent Q: What are the relations between Newton and gravity? \\
A: Newton discovered gravity. \\
A: Newton invented gravity. \hfill \break

\noindent Q: What are the relations between an electron and nucleus? \\
A: An electron revolves around the nucleus. \\
A: An electron is much smaller than the nucleus. \\
A: An electron attracts the nucleus. \hfill \break

\noindent Q: What are the relations between water and a pipe? \\
A: Water flows through the pipe. \\
A: Water passes through the pipe.

\subsubsection{Prompts used for baseline comparison}

After some experimentation with prompt engineering, we chose two variants of the prompt: 

\noindent {\it Q: Find an analogical mapping between the entities “eraser”, “paper” and “pencil” and the entities “keyboard”, “delete” and “screen”. \\ 
A: eraser:pencil:paper::delete:keyboard:screen} \\
\ \ \ \  \hphantom{}  \hspace{3.2cm}  {\it (or) \\
A: eraser -> delete, pencil -> keyboard, paper \\ -> screen}

\subsubsection{Possible leakage}

\begin{figure}
    \includegraphics[width=0.45\textwidth]{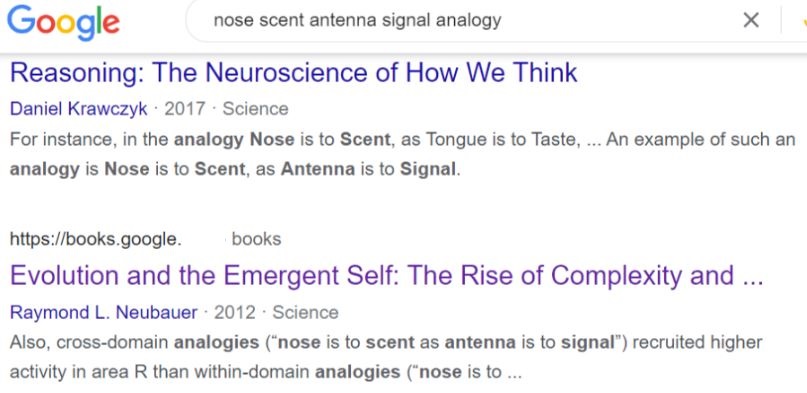} 
    \hrule
    
    \includegraphics[width=0.45\textwidth]{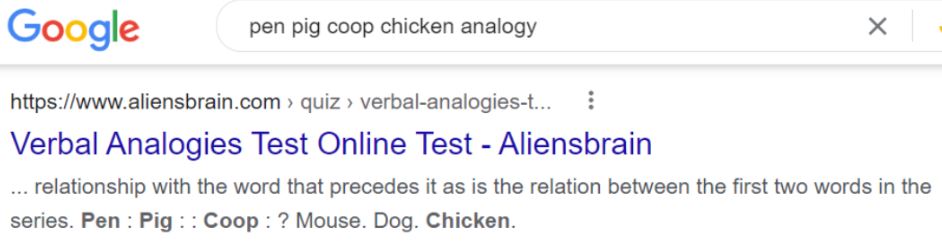}
    \caption{Looking for analogies from the original Green eval dataset online. 
    }
    \label{fig:noseAnalogy}
\end{figure}

Example answers for chosen analogies from Green eval dataset found via a simple web search can be found in Figure \ref{fig:noseAnalogy}

\label{app:gpt}

\subsection{Repository}
To ease the access and usage of our code we use Docker. Its main goal is to shift the cross-platform installation burden from the user to the developer. Unfortunately, we cannot share our Docker due to anonymity concerns (username). We will include it in the non-anonymized version. 

We provide a React based web interface, currently available only locally. This system is used to visualize the graphs created by the algorithm's mapping output. In addition, it visualizes the relations between entities, their similarity, and the clustering. This interface is useful for assisting in developing, debugging and understanding the algorithm's output. The demo is accessible using our repository\footref{ft:repoLink}. 
\label{app:docker}

\subsection{Experiments}
\begin{figure*}[ht]
    \centering
        \includegraphics[width=\textwidth]{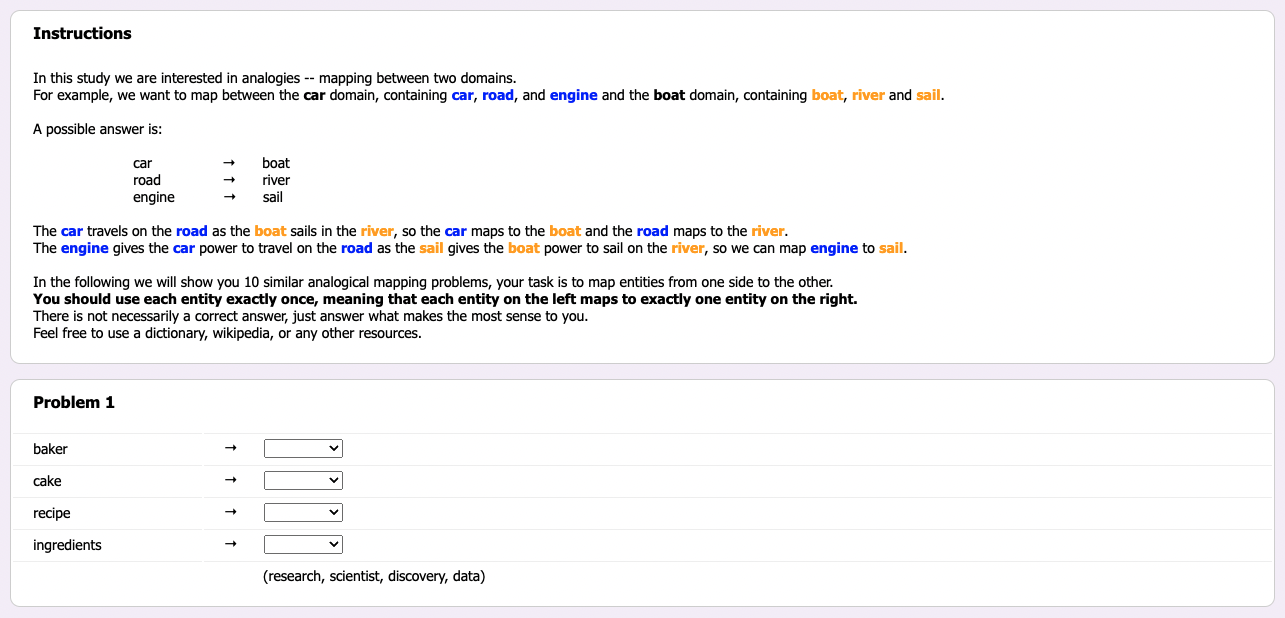}
        \caption{Closed-World Mapping: Experiment instructions with the first question.}
        \label{fig:experiment1}
\end{figure*}

\begin{figure*}[ht]
    \centering
        \includegraphics[width=\textwidth]{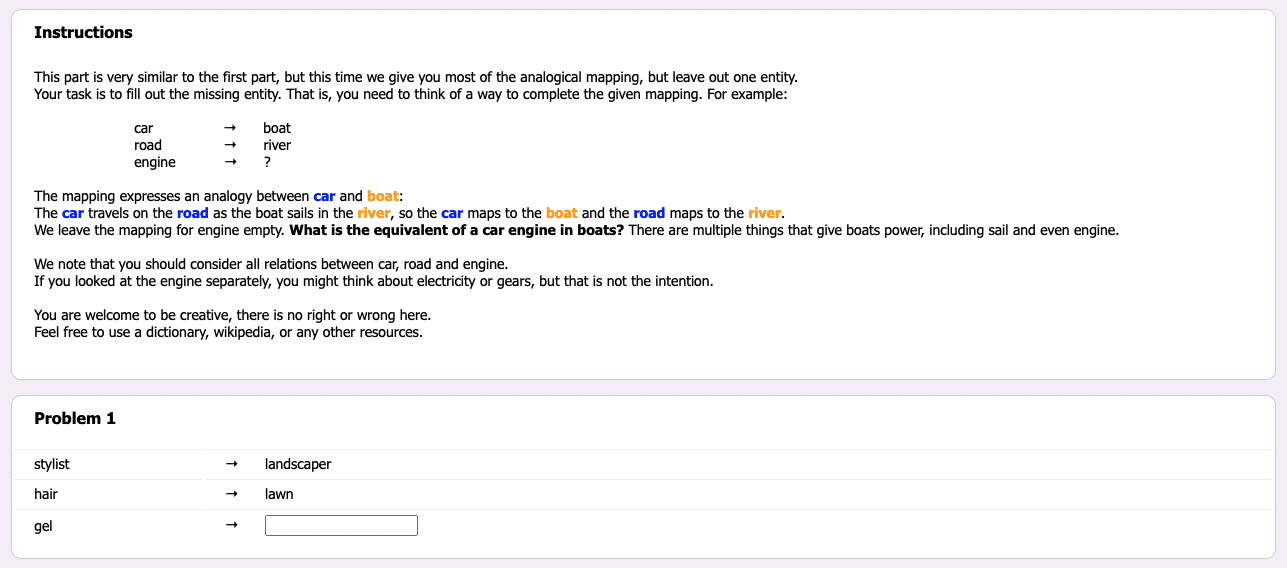}
        \caption{Open-World Entity Suggestion: Experiment instructions with the first question.}
        \label{fig:experiment2}
\end{figure*}

Snippets of the experimental setup (including instructions) can be found in Figures \ref{fig:experiment1}, \ref{fig:experiment2}. 

Table \ref{fig:part1problems} depicts the ten analogical proportion problems used in the \textit{structure mapping} experiment (closed-world mappings in Section \ref{sec:evaluation}). Accuracy denotes the percentage of human participants who mapped from \base to \targ correctly. Results show this task is non-trivial even for humans.

Table \ref{fig:part2problems} illustrates the experimental setup for the second phase of our experiment, in which participants received a solved mapping problem with one entity left out (open-World in Section \ref{sec:evaluation}). 

Table \ref{fig:part2example} contains all solved analogy problems used in the second phase of the experiment (entity suggestion, see open-World in Section \ref{sec:evaluation}). Participants were given with the complete mapping, but with a missing entity (as presented here). 

\begin{table*}[t]
    \begin{center}
        \begin{tabular}{c l@{\hskip 0.01in}c@{\hskip 0.2in}l @{\hskip 0.00001in}c}
            \hline
             & {\ \ \ \ \base} & \textbf{Mapping} & {\ \ \ \ \ \ \targ} & \shortstack{\textbf{Human Accuracy}\\\textbf{(Guess Level)}} \\      
            \hline
            \multirow{4}{*}{$\textbf{A1}$}& Baker & \multirow{4}{*}{$\rightarrow$} & Scientist & \multirow{4}{*}{\shortstack{$79.6\%$\\($4.2\%$)}}\\
            & Cake & & Discovery \\
            & Recipe & & Research \\
            & Ingredients & & Data \\
            
            \hline
            \multirow{3}{*}{$\textbf{A2}$}& Eraser & \multirow{3}{*}{$\rightarrow$} & Amnesia & \multirow{3}{*}{\shortstack{$71.7\%$\\($16.7\%$)}}\\
            & Pencil & & Memory \\
            & Paper & & Mind \\
            
            \hline
            \multirow{3}{*}{$\textbf{A3}$}& Jacket & \multirow{3}{*}{$\rightarrow$} & Wound & \multirow{3}{*}{\shortstack{$68.8\%$\\($16.7\%$)}}\\
            & Zipper & & Suture \\
            & Cold & & Infection \\
            
            \hline
            \multirow{3}{*}{$\textbf{A4}$}& Train & \multirow{3}{*}{$\rightarrow$} & Signal & \multirow{3}{*}{\shortstack{$74.0\%$\\($16.7\%$)}}\\
            & Track & & Wire \\
            & Steel & & Copper \\
            
            \hline
            \multirow{3}{*}{$\textbf{A5}$}& Thoughts & \multirow{3}{*}{$\rightarrow$} & Astronaut & \multirow{3}{*}{\shortstack{$53.9\%$\\($16.7\%$)}}\\
            & Brain & & Space \\
            & Neurons & & Stars \\
            
            \hline
            \multirow{5}{*}{$\textbf{A6}$}& Water & \multirow{5}{*}{$\rightarrow$} & Heat & \multirow{5}{*}{\shortstack{$35.5\%$\\($0.8\%$)}}\\
            & Pressure & & Temperature \\
            & Bucket & & Kettle \\
            & Pipe & & Iron \\
            & Rain & & Sun \\
            
            \hline
            \multirow{4}{*}{$\textbf{A7}$}& Waves & \multirow{4}{*}{$\rightarrow$} & Sounds & \multirow{4}{*}{\shortstack{$65.1\%$\\($4.2\%$)}}\\
            & Water & & Air \\
            & Shore & & Ear \\
            & Breakwater & & Earplugs \\
            
            \hline
            \multirow{4}{*}{$\textbf{A8}$}& Goal & \multirow{4}{*}{$\rightarrow$} & Basket & \multirow{4}{*}{\shortstack{$94.1\%$\\($4.2\%$)}}\\
            & Soccer & & Basketball \\
            & Grass & & Hardwood \\
            & Feet & & Hands \\
            
            \hline
            \multirow{3}{*}{$\textbf{A9}$}& Seeds & \multirow{3}{*}{$\rightarrow$} & Ideas & \multirow{3}{*}{\shortstack{$64.5\%$\\($16.7\%$)}}\\
            & Fruit & & Product \\
            & Bloom & & Success \\
            
            \hline
            \multirow{4}{*}{$\textbf{A10}$}& Morning & \multirow{4}{*}{$\rightarrow$} & Evening & \multirow{4}{*}{\shortstack{$95.1\%$\\($4.2\%$)}}\\
            & Breakfast & & Dinner \\
            & Start & & End \\
            & Coffee & & Wine \\
            
            \hline
        \end{tabular}
        \centering
        \caption{The ten analogical proportion problems used in the \textit{structure mapping} experiment. Accuracy denotes the percentage of human participants who mapped from \base to \targ correctly. Note that each row under the \base column is mapped to its \targ column. Problem's guess level appears in brackets below the accuracy. Results show this task is non-trivial even for humans.}
        \label{fig:part1problems}
    \end{center}
\end{table*}

\begin{table*}[ht!]
    \begin{center}
        \begin{tabular}{ l@{\hskip 0.2in}c@{\hskip 0.4in}l }
            \hline
            \ \ \ \ \ \base & \textbf{Mapping} & \ \ \ \targ \\ 
            \hline
            Electrons & $\rightarrow$ & Earth \\
            Electricity & $\rightarrow$ & Gravity \\
            Faraday & $\rightarrow$ & Newton \\
            Nucleus & $\rightarrow$ & ? \\
            \hline
        \end{tabular}
        \centering
        \caption{Solved mapping problem with one missing \targ entity. Participants instructed to fill in the missing entity.}
        \label{fig:part2example}
    \end{center}
\end{table*}

\begin{table*}[ht]
    \begin{center}
        \begin{tabular}{m{1.5em} m{3.5em} m{2em} m{4.5em} |>{\centering\arraybackslash}m{9em} |>{\centering\arraybackslash}m{9em}} 
            \hline
              &\ \ \ \ \ \base &  &\ \ \ \ \ \  \targ & \textbf{Algorithm} & \textbf{Humans} \\ 
             \hline
            
            \multirow{4}{*}{$\textbf{B1}$} & Answer & \multirow{4}{*}{$\rightarrow$} & Key & & \\
            & Logic & & Mechanism & &  \\
            & Riddle & & \textbf{?} & &  \\
            & & & & \makecell{Problem \\ \textbf{Lock} \\ Feedback} & \makecell{\textbf{Lock} (58.9\%) \\ Door (11.8\%) \\ Question (4.6\%)} \\
            \hline
            
            \multirow{5}{*}{$\textbf{B2}$} & Earth & \multirow{5}{*}{$\rightarrow$} & Electrons & & \\
            & Gravity & & Electricity & &  \\
            & Newton & & Faraday & &  \\
            & \textbf{?} & & Nucleus & & \\ & & & & \makecell{\textbf{Sun} \\ Moon \\ Mars} & \makecell{Earth's core (15.8\%) \\ Apple (13.2\%) \\ \textbf{Sun} (10.2\%)} \\
            \hline
            
            \multirow{4}{*}{$\textbf{B3}$} & Stylist & \multirow{4}{*}{$\rightarrow$} & Landscaper & & \\
            & Hair & & Lawn & &  \\
            & Gel & & \textbf{?} & &  \\
            & & & & \makecell{\textbf{Fertilizer} \\ Water \\ Lime} & \makecell{\textbf{Fertilizer} (29.3\%) \\ Lawn Mower (21.1\%) \\ Shears (10.2\%)} \\
            \hline

            \multirow{5}{*}{$\textbf{B4}$} & Chef & \multirow{5}{*}{$\rightarrow$} & Baker & & \\
            & Meal & & Cake & &  \\
            & Pan & & Oven & &  \\
            & Salt & & \textbf{?} & & \\ & & & & \makecell{Butter \\ \textbf{Sugar} \\ Onion} & \makecell{\textbf{Sugar} (63.5\%) \\ Flour (6.9\%) \\ Pepper (3.3\%)} \\
            \hline
            
            \multirow{4}{*}{$\textbf{B5}$} & Sun & \multirow{4}{*}{$\rightarrow$} & Rain & & \\
            & Summer & & Winter & &  \\
            & Sunscreen & & \textbf{?} & &  \\
            & & & & \makecell{\textbf{Umbrella} \\ Birds \\ Flooding} & \makecell{\textbf{Umbrella} (51.0\%) \\ Coat (20.7\%) \\ Cream (9.9\%)} \\
            \hline
            
        \end{tabular}
    \end{center}
    \caption{Examples used in the second phase of the experiment. Participants were given with the complete mapping, but with a missing entity (as presented here). The algorithm top three completions are sorted according to certainty. Humans' top three completions are sorted according to their frequency in the experiment (in brackets).}
    \label{fig:part2problems}
\end{table*}
\label{app:experiments}

\subsection{E-kar}
Table \ref{fig:ekar} shows an example of a problematic problem from E-KAR dataset. 

\begin{table}[t!]
    \begin{center}
        \begin{tabular}{ l@{\hskip 0.2in}c@{\hskip 0.4in}l }
            \hline
            \ \ \ \ \ \ \ \ \base & \textbf{Mapping} & \ \ \ \ \ \ \targ \\ 
            \hline
            Ice & $\rightarrow$ & Grass \\
            Fog & $\rightarrow$ & Tree \\
            \hline
        \end{tabular}
        \centering
        \caption{"ice" and "fog" are different forms of the same substance, and both "ice" and "fog" are natural objects.". "grass" and "tree" are both plants, and "grass" and "tree" are both natural objects.}
        \label{fig:ekar}
    \end{center}
\end{table}  

\label{app:ekar}

\end{document}